# Comparative Analysis of Machine Learning Approaches to Analyze and Predict the Covid-19 Outbreak

Muhammad Naeem[1], Jian Yu[2], Muhammad Aamir[1], Sajjad Ahmad Khan[3], Olayinka Adeleye[2], Zardad Khan[1]

[1]Department of Statistics, Abdul Wali Khan University Mardan, KP, Pakistan
[2]Department of Computer Science, Auckland University, New Zealand
[3]Department of Statistics, Islamia College University Peshawar, KP, Pakistan

Corresponding author:
Muhammad Aamir
Department of Statistics, Abdul Wali Khan University Mardan, 23200, KP, Pakistan
Email address: aamirkhan@awkum.edu.pk

## Abstract

**Background.** Forecasting the time of forthcoming pandemic reduces the impact of diseases by taking precautionary steps such as public health messaging and raising the consciousness of doctors. With the continuous and rapid increase in the cumulative incidence of COVID-19, statistical and outbreak prediction models including various machine learning (ML) models are being used by the research community to track and predict the trend of the epidemic, and also in developing appropriate strategies to combat and manage its spread.

**Methods.** In this paper, we present a comparative analysis of various ML approaches including Support Vector Machine, Random Forest, K-Nearest Neighbor and Artificial Neural Network in predicting the COVID-19 outbreak in the epidemiological domain. We first apply autoregressive distributed lag (ARDL) method to identify and model the short and the long-run relationships of the time-series COVID-19 datasets. That is, we determine the lags between a response variable and its respective explanatory time series variables as independent variables. Then, the resulting significant variables concerning their lags are used in the regression model selected by the ARDL for predicting and forecasting the trend of the epidemic.

**Results.** Statistical measures i.e., Root Mean Square Error (RMSE), Mean Absolute Error (MAE), and Mean Absolute Percentage Error (MAPE) are used for model accuracy. The values of MAPE for the best selected models for confirmed, recovered and deaths cases are 0.407, 0.094 and 0.124 respectively, which falls under the category of highly accurate forecasts. In addition, we computed fifteen days ahead forecast for the daily deaths, recover, and confirm patients and the cases fluctuated across time in all aspect. Besides, the results reveals the advantages of ML algorithms for supporting decision making of evolving short term policies.

# Introduction

The outbreak of the novel coronavirus disease in 2019 (COVID-19) has emerged as one of the most devastating respiratory disease since the 1918 HIN1 influenzas pandemic, infecting millions of people globally (Tuli et al. 2020). The cumulative incidence of the virus is continually and rapidly increasing globally. At the early stage of the outbreak, it is important to have a clear understanding of the disease transmission and its dynamic progression, so that relevant agencies and organizations can make informed-decisions and enforce appropriate control measures. Generally, capturing the transmission dynamics of a disease over time can provide insights into its progression, and show whether the outbreak control measures are effective and able to reduce the impact of the disease on a community (Kucharski et al. 2020).

Access to real-time data and effective application of outbreak prediction or forecasting models are central to obtaining insightful information regarding the transmission dynamics of the disease and its consequences. Moreover, every outbreak has its unique transmission characteristics that are different from the other outbreaks, which raises the question of how standards prediction models would perform in delivering accurate results. In addition, various factors including the number of known and unknown variables, differences in population/behavioural complexity in various geopolitical areas, and the variations in containment strategies increase the uncertainty of prediction models (Ardabili et al. 2020). As a result, it is challenging for standard epidemiological models such as Susceptible-Infected-Recovered (SIR) to provide reliable results for long-term predictions. Therefore, it is important to not only study the relationship between the components of the outbreak datasets but also evaluate the effectiveness of the common disease prediction models.

In recent months, there have been a handful of works that try to understand the spread of COVID-19, especially using statistical approaches. For instance, Kucharski et al. explored a combination of stochastic transmission model and four datasets that captured the daily number of new cases, the daily number of new internationally exported cases, the proportion of infected passengers on evacuation flight and the number of new confirmed cases, to estimate the transmission dynamics of the disease over some time (Kucharski et al. 2020).

In another study, a machine learning-based model is applied to analyse and predict the growth of COVID-19 (Tuli et al. 2020). The authors demonstrated the effectiveness of using iterative weighting for fitting Generalized Inverse Weibull distribution when developing a prediction solution. Lin et al., presented a conceptual model designed for the COVID-19 epidemic with consideration of individual behavioural responses and engagements with the government, including extension in holidays, restriction on travel, quarantine, and hospitalization (Lin et al. 2020). This work combined zoonotic transmission with the emigration pattern, and then estimate the future trends and the reporting proportion. The model gives promising insight into the trend of the COVID-19 outbreak, especially the impact of individual and government reactions or responses to the epidemic. The authors (Anastassopoulou et al. 2020) estimated the average values of the key epidemiological parameters including the per day case mortality, recovery ratios, and the basic reproduction number $R_0$ representing the average number of ancillary cases that results from the introduction of a single infectious case in an entirely susceptible population during the active period of the pandemic. The authors fit the dataset to the Susceptible-Infectious-Recovered-Dead (SIDR) model and attempted a three-week prediction of the dynamics of the outbreak at the epic centre. The estimated mean value of $R_0$ as calculated considering the period from the 11$^{th}$ January to 18$^{th}$ of January was found to be around 2.6 based on the official confirmed cases. The authors (Hu et al. 2020) proposed a machine learning approach to predict the magnitude, intervals, and completion period of the disease. The authors proposed an improved auto-encoder model to analyse the spread changing aspects of the epidemics then predict the definite cases. In the model, hidden variables are used to group the cities for probing the spread arrangement. By means of the many-step predicting, the expected errors of 6,7,8,9 and 10-step predicting remained 1.64%, 2.27%, 2.14%, 2.08%, 0.73%, correspondingly.

Autoregressive Distributed Lag (ARDL) is a flexible method to include independent series in dynamic regression models. ARDL models contain previous values of together response and explanatory variables series. They have been widely used in various domains including marketing, energy, epidemiology, agronomy and ecological studies (Huffaker & Fearne 2019). Over the years, many packages have been developed for ARDL. For example (Pesaran et al. 2001). The distributed lag model has a wide range of application i.e. cointegration study in which small and large-run relations between time series data. ARDL boundaries testing of (Pesaran et al. 2001), which is a common co-integration study technique founded on the distributive lag model and further research work in progress.

The other package developed by (Demirhan 2020) is `nardl` to use Distributed Lag Models (DLMs) in R. The package `nardl` focuses on the application of the nonlinear cointegrating ARDL model is developed by (Shin et al. 2014). The recent package `dynlm` takes a unique purpose to fit linear models via stabilizing time-series features (Zeileis & Zeileis 2019). In the current study, we will use the R programming and `dLagM` package that outfits the ARDL test method (Pesaran et al. 2001). Subsequently, `dLagM` uses lag orders, dataset, and overall method which make the prerequisite lags and changes for a definite models. One of the benefits of this approach is that the users are not required to specify the variation for the applied models. Which brings efficacy and value to researchers in various areas.

In this work, we present a comparative analysis of various machine learning approaches including Support Vector Machine (SVM), Random Forest (RF), K-Nearest Neighbor (KNN) and Artificial Neural Network in predicting the COVID-19 outbreak in the epidemiological domain. We aim to determine how well each of these approaches performs in predicting the confirmed and death cases and then compare their performances with each other. Particularly, we first apply ARDL method to identify and model the short and the long-run relationships of the time-series COVID-19 datasets (confirmed, recovered and death cases). That is, we determine the lags between a response variable and its respective explanatory time series variables as independent variables. Then, the resulting significant variables concerning their lags are used in the regression model selected by the ARDL model for predicting and forecasting the trend and dynamics of the COVID-19. We evaluated the models using relevant accuracy and error metrics including Root Mean Square Error (RMSE), Mean Absolute Error (MAE) and Mean Absolute Percentage Error (MAPE).

## Materials & Methods

**Data Source**

We conducted our study based on the publicly accessible data of daily deaths, recovered and confirmed cases 17549, 332062 and 694123 respectively reported for all over the world from 22$^{nd}$ January 2020 to 18$^{th}$ Jan 2021, (Fig. 1). The data is available in the online repository - GitHub (https://github.com/CSSEGISandData/COVID-19). We perform data processing including the conversion of data format from cumulative to daily basis. This repository is for COVID-19 visual dashboard operated by Johns Hopkins University Centre Systems Science and Engineering (JHU CSSE). They have aggregated data from sources like WHO, WorldoMeters, BNO News, and Washington State Department of Health and many more. The data have the number of confirmed cases, the recovered cases, and the death cases for the global. On this data, we attempted to forecast the key epidemiological parameters, i.e., the number of upcoming daily new confirmed cases, deaths, and recoveries. Though, the quantity of deaths, recovery, and confirmed cases of individuals is expected to be much higher along time. Therefore, we have similarly derived a correlation between these two variables and their past record (lags) by using the ARDL model.

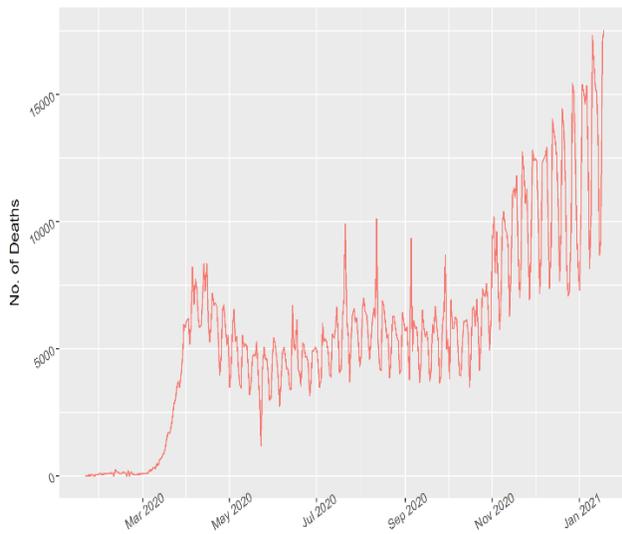
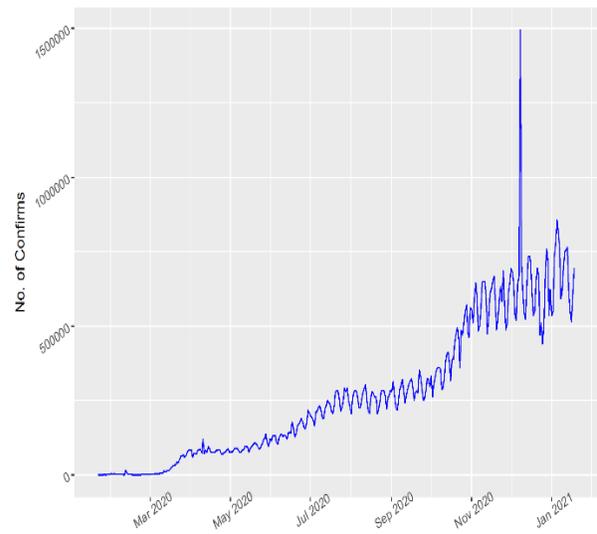

(a)　　　　　　　　　　　　　　　　　　　(b)

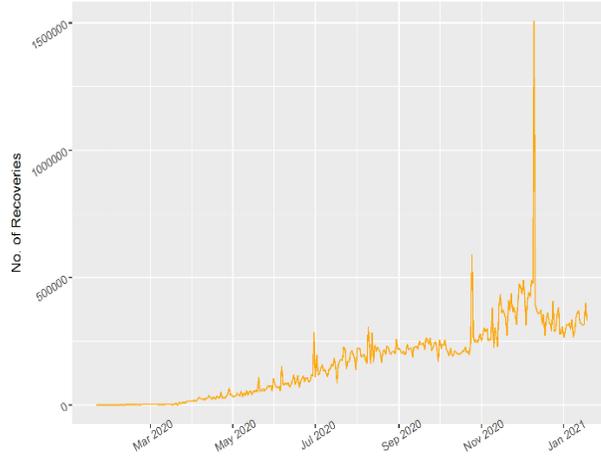

(c)

**Fig. 1.** Plots of the daily Deaths, Recovered and Confirmed COVID-19 outbreak (a) Deaths (Minimum = 1, Maximum = 17549, Median value = 5466) (b) Recoveries (Minimum = 2, Maximum = 1507012, Median value = 168554) and (c) Confirmed cases (Minimum = 98, Maximum = 1495214, Median value = 225212)

**Autoregressive Distributive Lag Models**

The ARDL models are used between regressed series and $k$ number of regressors series in regression analysis. In this study, we used `tseries, timeseries, zoo` and `window` packages for the data. In the same way, `dLagM` package in R for ARDL model. An orders $p$ and $q$ of the ARDL lag model are denoted by ARDL $(p, q)$, which has independent $p$ lags series and dependent $q$ lags series. If there is only one independent series, the dependent lag series makes the model autoregressive. The numeral of $p^{th}$ independent lag series is denoted by $p_j, j = 1, \ldots, n$ denotes daily recover and confirm cases, the $q^{th}$ lags of dependent variable series are shown by $q_i$, where $i = 0, 1, \ldots, m$.

The ARDL model can be expressed as:

$$y_t = \alpha_0 + \beta_1 y_{t-1} + \beta_2 y_{t-2} + \cdots, \beta_j y_{t-i} + \gamma_1 x_t + \gamma_2 x_{t-1} + \cdots, \gamma_i x_{t-j} +$$
$$\delta_1 w_t + \delta_2 w_{t-1} + \cdots, \delta_i w_{t-j} + \varepsilon_t \qquad (1)$$

Where $y_t$ denotes the number of daily deaths at time $t$. $\alpha_0$ represent the intercept term. In the same way, $\beta_1 y_{t-1} + \beta_2 y_{t-2} + \cdots, \beta_q y_{t-i}$ denotes the $q^{th}$ autoregressive lag order of the model of the dependent variable. The two independent variables "recover cases" and "confirm cases" are denoted by $x_t$ and $w_t$ respectively. Whereas $\gamma_1 x_t + \gamma_2 x_{t-1} + \cdots, \gamma_i x_{t-j}$ and $\delta_1 w_t + \delta_2 w_{t-1} + \cdots, \delta_i w_{t-j}$ represent the lags order of $x_t$ and $w_t$ respectively. The parameters $\beta, \gamma \: and \: \delta$ denoted coefficients of death, recover, and confirm cases, respectively, while $\varepsilon_t$ denotes the error term. Eq.1 can be further simplified and presented in (Eq. 2):

$$y_t = \alpha_0 + \sum_{i=1}^{m} \beta_i y_{t-i} + \sum_{j=0}^{n} \gamma_j x_{t-j} + \sum_{j=0}^{n} \delta_j w_{t-j} + \varepsilon_t \qquad (2)$$

The number of death, confirm, and recover cases of people is likely to be much higher with time. Therefore, the ARDL model for recovered cases $x_t$ and confirmed cases $w_t$ is shown in (Eq.3).

$$x_t = \theta_0 + \gamma_1 x_{t-1} + \cdots, \gamma_i x_{t-j} + \delta_1 w_t + \delta_2 w_{t-1} + \cdots, \delta_i w_{t-j} + \varepsilon_t \qquad (3)$$

Similarly, the ARDL model for confirmed and recovered cases is shown in (Eq.4)

$$w_t = \vartheta_0 + \delta_1 w_t + \delta_2 w_{t-1} + \cdots, \delta_i w_{t-j} + \gamma_1 x_{t-1} + \cdots, \gamma_i x_{t-j} + \varepsilon_t \qquad (4)$$

There are different criteria's used to select an optimal lag length selection. The authors in (Chandio et al. 2020) use Akaike Information Criterion (AIC) and the authors in (Gayawan & Ipinyomi 2009) compare AIC, SIC and adj-R square to select the optimal lag length. We use adj-R square and parsimony model criteria to select an optimal number of the lag length in this study. It makes the call to the function easier when the number of lags order are the same, however, when the number of lags order is different from dependent and every independent sequence, we use the argument *remove*. It will remove the lags that are not contributed to the model. Once the ARDL model specifies the significant coefficients of the dependent variable and independent variables, the models including the RF, SVM, KNN, and ANN are used to assess the accuracy and error rate of these models. We utilized RF (Biau & Scornet 2016), SVM (Liang et al. 2018), KNN (Martínez et al. 2019) and ANN (Hu et al. 2018) time series models were applied to predict the COVID-19. To overcome the overfitting problem, we use 80% training and 20% testing parts, respectively. Random forest is one of the best learning algorithms and it requires a bit parameter tuning. In the current study, `randomForest`, `forecast`, `caret`, `tiyverse`, `tsibble` and `purr` package are used for RF. the `ntree` is 500, `mtry` is p/3, where p is the number of features, `sampsize` is 70% and `type` is "regression" utilized in the function. The other parameters are kept as default.

Generally, in time series analysis, Support Vector Regression (SVR) is used. In this study, we use `e1071` library, the parameters *cost*=$10^2$, *gamma*($\gamma$) = 0.1, and the *insensitivity* ($\epsilon$) = 0.3 respectively. But the function we called SVM and it automatically chooses SVR or SVM when it detects continuous/categorical response of the data respectively. In SVM, various kernel functions are used to develop the input space into a feature space with a complex dimension. Like Gaussian Radial Basis (GRBF), Sigmoid, polynomial, etc. are some kernel functions. For SVM, we use Radial Basis Kernels (RBF) $k_\gamma(y_i, y_j) = \exp(-\gamma \|y_i - y_j\|)^2$. In the SVM model, using RBF kernels it is necessary to tune model parameters to find an optimal value of the parameters and reducing the overfitting problem. So, we use the grid search method of tenfold cross-validation on the training part and testing part and their results are averaged.

k-nearest neighbor (k-NN) predicts the response variable based on the nearest training points. In this study, we use `caret` package for k-nearest neighbor Regression in R. It uses a training dataset in its place of learning a discriminative function from the training data. k-NN is used both for classification and regression problems. There are various techniques use to improve model accuracy. Such as maximum percentage accuracy graph, Elbow method, for loops to select an optimal value of k. Generally, the square root of n is used, and we utilized $\sqrt{n}$.

ANN is a mathematical tool and has been generally used for classification and forecasting problems properly that contain predictors (input) and response (output) layers, and a hidden layer. A combination of different hidden layers is used to choose a better MLP architecture network. It is the hidden layers in ANN models that play an important role in many successful applications of neural networks. In the current study, we use `neuralnet` package for ANN. The parameters, the *algorithm, threshold,* and *linear.output* is 'backprop', 0.01, TRUE and the other parameters are kept as default, respectively. ANN model is widely used in the economic and financial studies (Huang et al. 2007; Qi 1996). The number of hidden layers depends upon the nature of the problem. The authors in (Zhang et al. 1998) used two hidden layers and finds better model prediction accuracy. In the same way, the authors in (Xu et al. 2020) used $(2 \times k + 1)$, where $k$ is the number of predictors (inputs). For an optimal result of ANN, usually, trial and error method is used in determining the number of hidden nodes i.e. searching the architecture having the smallest MAPE among the models (Güler & Übeyli 2005). We use 4 hidden layers and 8 neurons in the hidden layers for daily death cases using trial and error procedure and 10,000 times iteration. In the same way, we use 2 hidden layers and 4 number of neurons in the hidden layers for daily recover cases.

**Forecast Evaluation Criterions**

In this study, as the response variable is continuous, therefore, the forecasting capacity of different machine learning approaches are evaluated by using five different criterions including mean error (ME), RMSE, Mean Absolute Error (MAE), Mean Percentage Error (MPE) and MAPE and presented in (Tab. 1). Where n represents the total number of prediction on training and testing parts respectively, $Y_t$ and $\hat{Y}_t$ representing the observed and predicted values, respectively.

**Tab. 1.** Forecasting evaluation measurement tools

| Criterion | Formula |
|---|---|
| Mean error | $ME = \frac{1}{n}\sum_{t=1}^{n}(Y_t - \hat{Y}_t)$ |
| Root mean square error | $RMSE = \sqrt{\frac{1}{n}\sum_{t=1}^{n}(\hat{Y}_t - Y_t)^2}$ |
| Mean absolute error | $MAE = \frac{1}{n}\sum_{t=1}^{n}|\hat{Y}_t - Y_t|$ |
| Mean percentage error | $MPE = \frac{1}{n}\sum_{t=1}^{n}(\frac{\hat{Y}_t - Y_t}{Y_t}) * 100$ |
| Mean absolute percentage error | $MAPE = \frac{1}{n}\sum_{t=1}^{n}|\frac{\hat{Y}_t - Y_t}{Y_t}| * 100$ |

# Results

A total of three data sets of COVID-19 (confirm, recover and death) are used to evaluate the performance of the different ML approaches and suggested the best model for forecasting the COVID-19 outbreak. All data sets consisting of the world daily confirm, recover and deaths cases. Every time series divided into training and testing sets of observations. The original data divided into 80% training and 20% testing parts and the first 80% of the total observations in every time series used as a training set whereas the rest 20% used as the testing set. To overcome the overfitting problem, we use 10-fold cross-validation for each of the models and then their results are averaged. In addition, we also used prediction accuracy for training parts. Each time series containing a total of 366 observations spanning (22 January 2020, to 18 Jan 2021), the first 252 observations spanning (22 January 2020, to 07 Nov 2020) belong to the training series and the rest 74 observations spanning (08 Nov 2020, to 18 Jan 2021) part of the testing series.

We use death, recover, and confirm cases from the COVID-19 dataset. The COVID-19 dataset is loaded into the R package environment, and then, we fit ARDL model to the Daily Deaths series $y_t$ with recover $R_t$ and confirm $C_t$ cases. We choose $p_1 = 3$, $p_2 = 1$, $and\ q = 1$ using adj-R square and parsimony of the model. The insignificant variables are removed and fit the ARDL model. The results obtained from the ARDL model are presented in (Tab.2).

**Tab. 2.** Summary of ARDL model 1 for Daily Deaths of COVID-19

| Coefficients | Estimate | Std. Error | t value | P-value |
|---|---|---|---|---|
| (Intercept) | 481.82 | 121.23 | 3.97 | 8.97E-05*** |
| Ct.t | 0.017 | 0.002 | 7.36 | 1.99E-12*** |
| Ct.1 | -0.011 | 0.003 | -3.40 | 0.000761*** |
| Ct.2 | -0.006 | 0.002 | -2.72 | 0.006785** |
| Rt.t | 0.003 | 0.001 | 2.38 | 0.017842* |
| Yt.1 | 0.811 | 0.033 | 24.23 | 5.44E-71*** |

'***' Significant at 1%, '**' Significant at 5% , '*' Significant at 10%
Residual standard error:   998.7
Multiple R-squared:   0.8541,          Adjusted R-squared:  0.8520
F-statistic:          284.6,           P-value:          < 2.2e-16

The coefficient related to confirm cases $C_t$ and its first lag are highly significant at 1% level and 5% level, respectively. Similarly, first lag of the response variable $y_t$ (daily deaths of COVID-19), are significant at the 1% level. In addition, the coefficient of recover cases (first and second lags) are also significant at 1% level and 5% level, respectively. Overall, the model is highly significant at the 1% level with a p-value smaller than 2.2e-16 with the adjusted R-squared equal to 85.2%. The fitted model can be written as:

$$y_t(Daily\ Deaths) = 481.82 + 0.811 y_{t-1} + 0.003 x_{t-1} + 0.017 w_t - 0.011 w_{t-1} - 0.006 w_{t-2} + \varepsilon_t \tag{5}$$

In the second scenario, we examine the relationship between the number of recover cases and confirm cases. We fit the ARDL model for recover cases $x_t$ of COVID-19 series with confirm $w_t$ cases. We take $p_1 = 4$, $and\ q = 3$ using adj-R square and parsimony of the model and fitting the ARDL model to the datasets. The results obtained from the ARDL model are presented in (Tab. 3).

**Tab. 3.** Summary of final ARDL model 2 for confirm and recover of Covid-19

| Coefficients | Estimate | Std. Error | t value | P-value |
|---|---|---|---|---|
| (Intercept) | 6001.03 | 5580.45 | 1.07 | 0.28294 |
| Rt.t | 0.153 | 0.049 | 3.11 | 0.00200** |
| Rt.1 | 0.124 | 0.062 | 1.99 | 0.04646* |
| Rt.2 | -0.140 | 0.060 | -2.30 | 0.02140* |
| Rt.3 | 0.093 | 0.049 | 1.91 | 0.05653· |
| Ct.1 | 0.695 | 0.043 | 15.99 | 1.15E-43*** |
| Ct.3 | -0.170 | 0.072 | -2.36 | 0.018714* |
| Ct.4 | 0.320 | 0.065 | 4.86 | 1.72E-06*** |

'***' Significant at 1%,  '**' Significant at 5% , '*' Significant at 10%

Residual standard error: 67560

Multiple R-squared:  0.9171,               Adjusted R-squared:  0.9155

F-statistic: 556.5,                         P-value:  0.0000012

Tab. 3 shows the summary of the ARDL model, the confirm cases recorded in the current day, and the third and fourth days. The daily recover cases of current, one day, two days and three days before have a significant impact on the number of daily recover cases from the COVID-19 on that particular day. The model is significant at the 1% level ($P < 0.0000012$), the adjusted R-squared value is 91.55%. The fitted model can be written as:

$x_t(Confirm) = 6001.03 + 0.695 x_{t-1} - 0.170 x_{t-3} - 0.320 x_{t-4} + 0.153 w_t + 0.124 w_{t-1} - 0.140 w_{t-2} + 0.093 w_{t-3} + \varepsilon_t$ (6)

**Tab. 4.** Summary of final ARDL model 3 for confirm and recover of Covid-19

| Coefficients | Estimate | Std. Error | t value | P-value |
|---|---|---|---|---|
| (Intercept) | 5992.99 | 4710.743 | 1.272196 | 0.204141 |
| Ct.t | 0.125 | 0.043208 | 2.902152 | 0.003939* |
| Ct.1 | -0.225 | 0.052399 | -4.30755 | 2.14E-05*** |
| Ct.2 | 0.777 | 0.054089 | 14.37865 | 3.19E-37*** |
| Ct.3 | -0.451 | 0.050888 | -8.88207 | 3.39E-17*** |
| Rt.1 | 0.389 | 0.048003 | 8.105144 | 8.73E-15*** |
| Rt.2 | 0.088 | 0.042017 | 2.094582 | 0.036921* |
| Rt.3 | 0.127 | 0.041329 | 3.078196 | 0.002246** |

'***' Significant at 1%,  '**' Significant at 5% , '*' Significant at 10%

Residual standard error: 57220

Multiple R-squared:  0.8568,               Adjusted R-squared:  0.8539

| F-statistic: 301.7, | P-value: 0.0000031 |
|---|---|

Tab. 4 shows the summary of the ARDL model, the confirm cases recorded in the current, first, second and the third lags. The daily confirm cases of the first lags have a significant impact on the number of daily confirm cases from the COVID-19 on first lag, second lag and the number of confirm cases. The model is significant at the 1% level ($P < 0.0000031$), the adjusted R-squared value is 85.39%. The fitted model can be written as:

$$w_t(Recovered) = 5992.99 + 0.389w_{t-1} + 0.088w_{t-2} + 0.127w_{t-3} + 0.125x_t - 0.225x_{t-1} + 0.777x_{t-2} - 0.451x_{t-3} + \varepsilon_t \qquad (7)$$

We evaluate models including RF, SVM, KNN, and ANN to compare their performance using various accuracy metrics including ME, RMSE, MAE, MPE and MAPE. These metrics provide different perspectives to assess predicting models. The first three are the absolute performance measures while the fourth and fifth are relative performance measures. The training sample is used to estimate the parameters for specific model architecture. The testing set is then used to select the best model among all models considered. Tab. 5 summarizes the RF, SVM, KNN, and ANN forecasting accuracy measures for the training set of COVID-19 daily deaths data.

**Tab. 5.** Forecasting accuracy measures of all models for daily deaths of training data

| Method | Error Measurement Tools | | | | |
|---|---|---|---|---|---|
|  | ME | RMSE | MAE | MPE | MAPE |
| RF | 88.22 | 7011.30 | 3920.83 | -6.99 | 8.94 |
| SVM | 349.01 | 10432.35 | 8890.14 | -139.11 | 131.19 |
| KNN | 112.93 | 13343.31 | 8202.91 | -4.01 | 8.07 |
| ANN | **4.21** | **4.01** | **2.11** | **-0.0310** | **0.020** |

In (Tab. 5), the values of ME for RF, SVM, KNN, and ANN models are 88.22, 349.01, 112.93 and 4.21 respectively. This reveals that RF shows the lowest value (the best) among the other methods. Similarly, the RMSE values of RF, SVM, KNN, and ANN are 7011.30, 10432.35, 13343.31 and 4.01, respectively and show that the ANN achieved better performance compare to the other methods. Moreover, the MAE values for RF, SVM, KNN, and ANN models are 3920.83, 8890.14, 8202.91 and 2.11, respectively. While the values of MPE for RF, SVM, KNN, and ANN models are -6.99, -139.11, -4.01 and -0.0310 respectively. The ANN achieved better performance compare to the other methods respectively. Similarly, the values of MAPE of the RF, SVM, KNN, and ANN models are 8.94, 131.19, 8.07 and 0.020, respectively. Thus, the value of the MAPE for ANN is less than 1 which indicates that the selected model fall in the range of perfect model (Gao et al. 2019). We highlighted the results for ANN model indicating the smallest value among all models. In the most, of the cases, the ANN method shows significant performance compare to the rest of the method's base on training parts.

**Tab. 6.** Forecasting accuracy measures of all models for daily deaths of testing data

| Method | Error Measurement Tools | | | | |
|---|---|---|---|---|---|
| | ME | RMSE | MAE | MPE | MAPE |
| RF | 3474.03 | 4313.07 | 3510.00 | 27.66 | 28.27 |
| SVM | 3977.75 | 5255.87 | 3981.07 | 31.28 | 31.33 |
| KNN | 2963.88 | 3882.66 | 3143.84 | 22.79 | 25.53 |
| ANN | **17.49** | **36.35** | **18.03** | **0.117** | **0.124** |

In (Tab. 6), the values of ME for RF, SVM, KNN, and ANN models are 3474.03, 3977.75, 2963.88 and 17.49 respectively. The results indicate that ANN shows the lowest value among the other methods. Similarly, the RMSE values of RF, SVM, KNN, and ANN are 4313.07, 5255.87, 3882.66 and 36.35, respectively and show that ANN achieved better performance compare to the other methods. Moreover, the MAE values for RF, SVM, KNN, and ANN models are 3510.00, 3981.07, 3143.84 and 18.03, respectively. This shows ANN is better as compared to the other methods. While the values of MPE for RF, SVM, KNN, and ANN models are 27.66, 31.28, 22.79 and 0.117, respectively. Similarly, the values of MAPE of the RF, SVM, KNN, and ANN models are 28.27, 31.33, 25.53 and 0.124, respectively. Thus, the value of the MAPE for ANN is less than 1 which indicates that the selected model fall in the range of perfect model (Gao et al. 2019). We highlighted the results for ANN model indicating the smallest value among all models. The ANN method shows significant performance compares to the rest of the method's base on 20% testing parts in most of the cases. Fig. 2 shows the plot of the forecasting accuracy measures for the models.

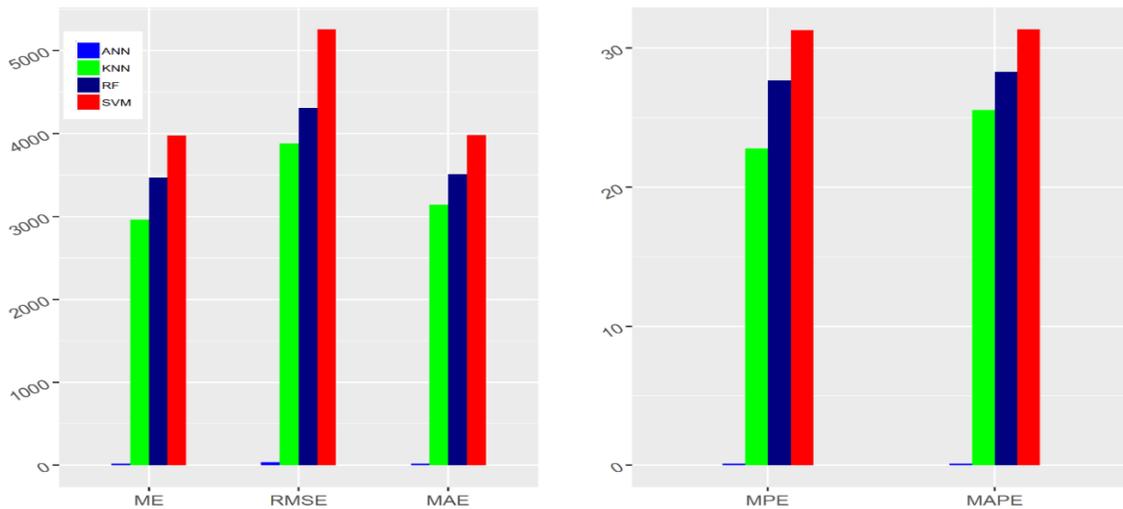

**Fig. 2.** Plot of the forecasting accuracy measures of different models for daily deaths. The best model is ANN their ME = 17.49, RMSE = 36.35, MAE = 18.03, MPE = 0.117 and MAPE = 0.124. The MAPE value is in the range of 0 to 1 which falls under the category of highly accurate forecasts (Aamir et al. 2018; Xu et al. 2020)

It is clear from the above plot that on the average, ANN is the best model for forecasting the daily deaths of COVID-19 outbreak. Tab. 7 summarizes the RF, SVM, KNN, and ANN forecasting accuracy measures of the COVID-19 confirm patient's on the training dataset.

**Tab. 7.** Forecasting accuracy measures of all models for daily confirm patients on training data

| Method | Error Measurement Tools | | | | |
|---|---|---|---|---|---|
| | ME | RMSE | MAE | MPE | MAPE |
| RF | 95.49 | 7085.34 | 4015.50 | -6.90 | 9.46 |
| SVM | 449.01 | 10823.55 | 8979.00 | -144.08 | 150.19 |
| KNN | 138.93 | 13971.41 | 8303.31 | -3.96 | 10.46 |
| ANN | **-0.0005** | **12.33** | **8.29** | **-0.056** | **0.0662** |

In (Tab. 7), the values of ME for RF, SVM, KNN, and ANN models are 95.49, 449.01, 138.93 and -0.0005, respectively. KNN has the lowest (the best) value among the other methods with the highest accuracy. Similarly, the RMSE values of RF, SVM, KNN, and ANN models are 7085.34, 10823.55, 13971.41 and 12.33, respectively. The ANN has shown the lowest RMSE value as compared to the rest of the methods. The MAE values of RF, SVM, KNN, and ANN models are 4015.50, 8979.00, 8303.31 and 8.29, respectively. The MAE value indicates that ANN has the smallest value among the other methods. While the values of MPE for RF, SVM, KNN, and ANN models are -6.90, -144.08, -3.96 and -0.056, respectively. Similarly, the values of MAPE of the ANN, SVM, RF and KNN models are 9.46, 150.19, 10.46 and 0.0662, respectively. Thus, the value of the MAPE for RF is in the range of 1 to 10 which revealed that the selected model falls in the category very good model. Overall, the ANN method achieved significant performance better than the other methods based on training parts. This indicates that ANN results are more consistent to RF, SVM, and KNN.

**Tab. 8.** Forecasting accuracy measures of all models for daily confirm patients on testing data

| Method | Error Measurement Tools | | | | |
|---|---|---|---|---|---|
| | ME | RMSE | MAE | MPE | MAPE |
| RF | 6513.24 | 14211.80 | 8451.63 | 7.67 | 11.55 |
| SVM | 15239.30 | 20942.20 | 15272.70 | 21.82 | 21.89 |
| KNN | 7859.90 | 15198.20 | 9790.63 | 9.65 | 13.58 |
| ANN | **1219.66** | **890.66** | **122.69** | **0.094** | **0.094** |

In (Tab. 8), the values of ME for RF, SVM, KNN, and ANN models 6513.24, 15239.30, 7859.90 and 1219.66, respectively. ANN has the lowest (the best) value among the other methods with highest accuracy. Similarly, the RMSE values of RF, SVM, KNN, and ANN models are 14211.80, 20942.20, 15198.20 and 890.66, respectively. The ANN has shown the lowest RMSE value as compared to the rest of the methods. The MAE values of RF, SVM, KNN, and ANN models are 8451.63, 15272.70, 9790.63 and 122.69, respectively. The MAE value indicates that ANN has smallest value among the other methods. While the values of MPE for RF, SVM, KNN, and ANN models are 7.67, 21.82, 9.65 and 0.094, respectively. The MPE value reveals that RF has smallest value among the other methods. Similarly, the values of MAPE of the ANN, SVM, RF and KNN models are 11.55, 21.89, 13.58 and 0.0942, respectively. Thus, the value of the MAPE for ANN is in the range of 1 to 10 which revealed that the selected model falls in the category very good model. On average, the ANN method achieved significant performance better than the other methods based on 20% testing parts. This indicates that ANN results are more consistent to RF, SVM, and KNN. Fig. 3 shows the plot of the forecasting accuracy measures for different models.

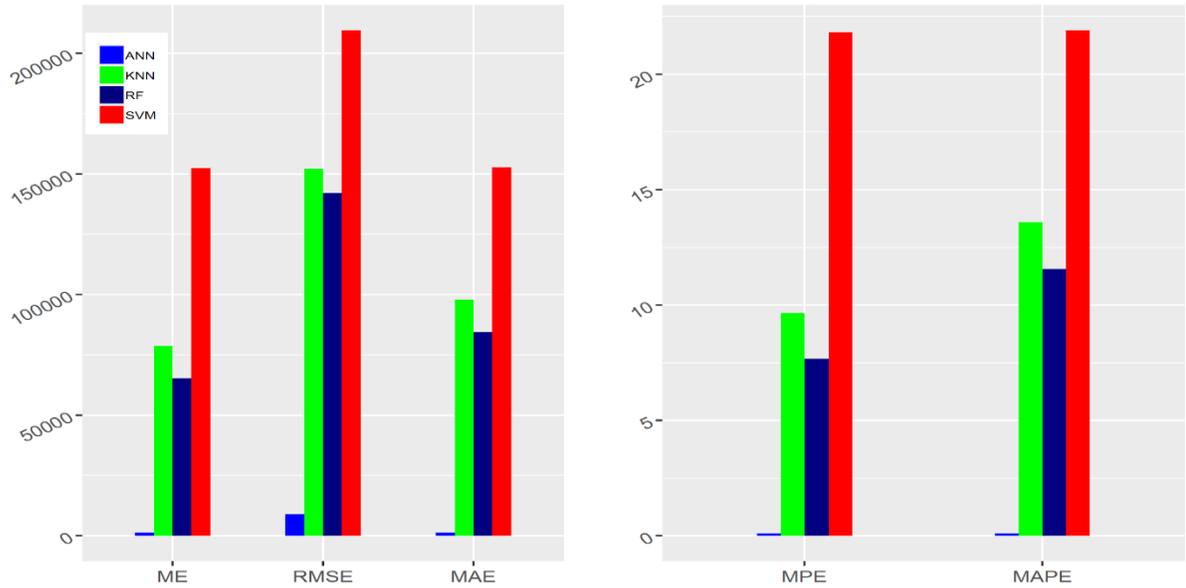

**Fig. 3.** Plot of the forecasting accuracy measures of different models for confirm cases. The best model is ANN their ME = 1219.66, RMSE = 890.66, MAE = 122.69, MPE = 0.0941 and MAPE = 0.0942. The MAPE value is in the range of 0 to 1 which falls under the category of highly accurate forecasts.

**Tab. 9.** Forecasting accuracy measures of all models for daily recover patients on training data

| Method | Error Measurement Tools | | | | |
|---|---|---|---|---|---|
| | ME | RMSE | MAE | MPE | MAPE |
| RF | 354.19 | 9979.71 | 3301.26 | -23.62 | 26.52 |
| SVM | -1936.20 | 16565.45 | 5981.51 | -2482.15 | 2484.31 |
| KNN | 2233.49 | 21385.82 | 8336.57 | -30.99 | 40.20 |
| ANN | **-0.0008** | **2.78** | **2.20** | **-0.2965** | **0.3022** |

In (Tab. 9), the values of ME for RF, SVM, KNN, and ANN models are 2.27, 449.01, 138.93 and 2.43, respectively. KNN has the lowest (the best) value among the other methods with highest accuracy. Similarly, the RMSE values of RF, SVM, KNN, and ANN models are 7119.56, 10823.55, 13971.41 and 2.14, respectively. The ANN has shown the lowest RMSE value as compared to the rest of the methods. The MAE values of RF, SVM, KNN, and ANN models are 4106.04, 8979.00, 8303.31 and 1.21, respectively. The MAE value indicates that ANN has smallest value among the other methods. While the values of MPE for RF, SVM, KNN, and ANN models are -7.28, -144.08, -3.96 and 0.0003, respectively. The MPE value reveals that ANN has smallest value among the other methods. Similarly, the values of MAPE of the ANN, SVM, RF and KNN models are 9.88, 150.19, 10.46 and 0.0029, respectively. Thus, the value of the MAPE for ANN is in the range of 1 to 10 which revealed that the selected model falls in the category very good model. On average, the ANN method achieved significant performance better than the other methods based on 20% testing parts. This indicates that ANN results are more consistent to RF, SVM, and KNN. Fig. 3 shows the plot of the forecasting accuracy measures for different models.

**Tab. 10.** Forecasting accuracy measures of all models for daily recover patients on testing data

| Method | Error Measurement Tools | | | | |
|---|---|---|---|---|---|
| | ME | RMSE | MAE | MPE | MAPE |
| RF | 7021.45 | 15078.57 | 7486.51 | 14.80 | 16.68 |
| SVM | 9729.33 | 18611.90 | 10126.07 | 21.27 | 22.92 |
| KNN | 8728.45 | 16763.27 | 9164.31 | 18.92 | 20.74 |
| ANN | **489.57** | **4384.81** | **538.98** | **0.2828** | **0.4070** |

In (Tab. 10), the values of ME for RF, SVM, KNN, and ANN models are 7021.45, 9729.33. 8728.45 and 489.57, respectively. ANN has the lowest (the best) value among the other methods with highest accuracy. Similarly, the RMSE values of RF, SVM, KNN, and ANN models are 15078.57, 18611.90, 16763.27 and 4384.81, respectively. The ANN has shown the lowest RMSE value as compared to the rest of the methods. The MAE values of RF, SVM, KNN, and ANN models are 7486.51,10126.07, 9164.31 and 538.98, respectively. The MAE value indicates that ANN has smallest value among the other methods. While the values of MPE for RF, SVM, KNN, and ANN models are 14.80, 21.27, 18.92 and 0.2828, respectively. The MPE value reveals that ANN has smallest value among the other methods. Similarly, the values of MAPE of the ANN, SVM, RF and KNN models are 16.68, 22.92, 20.74 and 0.4070, respectively. Thus, the value of the MAPE for ANN is in the range of 1 to 10 which revealed that the selected model falls in the category very good model. On average, the ANN method achieved significant performance better than the other methods based on 20% testing parts. This indicates that ANN results are more consistent to RF, SVM, and KNN. Fig. 4 shows the plot of the forecasting accuracy measures for different models.

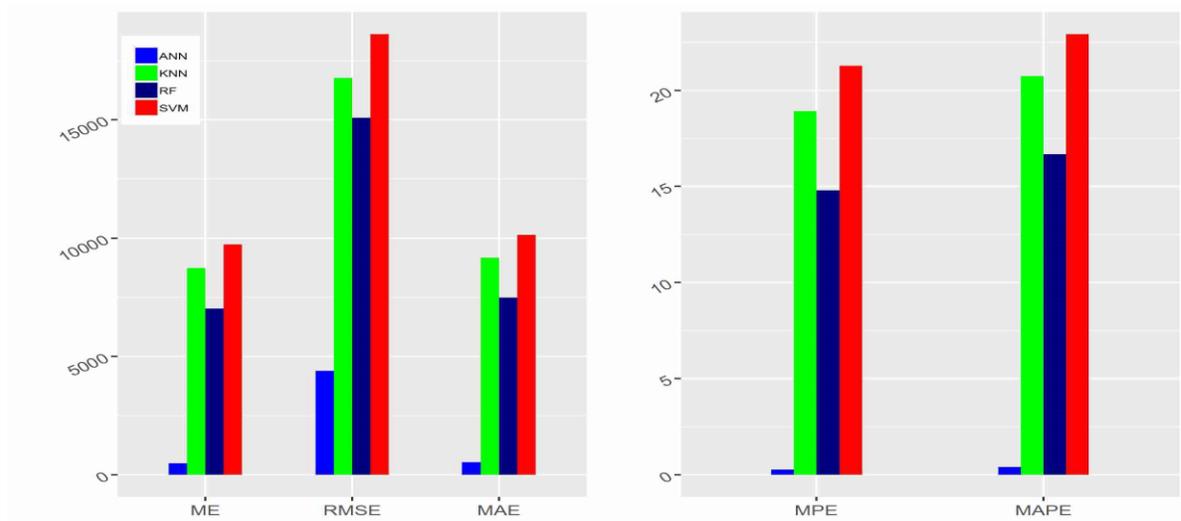

**Fig. 4.** The plot of the forecasting accuracy measures of different models for recover cases. The best model is ANN their ME = 489.57, RMSE = 4384.81, MAE = 538.98, MPE = 0.2828 and MAPE = 0.4070. The MAPE value is in the range of 0 to 1 which falls under the category of highly accurate forecasts.

## Discussion

The performance of the neural network model can be assessed once trained the network employing the performance function as a prediction. All the methods are capable of capturing the pattern of the data effectively. Moreover, ANN performed well and almost capture the whole pattern of the testing part of the data when compared to RF, SVM, and KNN methods. Fig.3 shown the prediction accuracy of the number of daily Covid-19 recovered cases of RF, SVM, KNN, and ANN methods. The world daily deaths original testing data of COVID-19 and the forecasted data for RF, SVM, KNN and ANN models are plotted in (Fig.5).

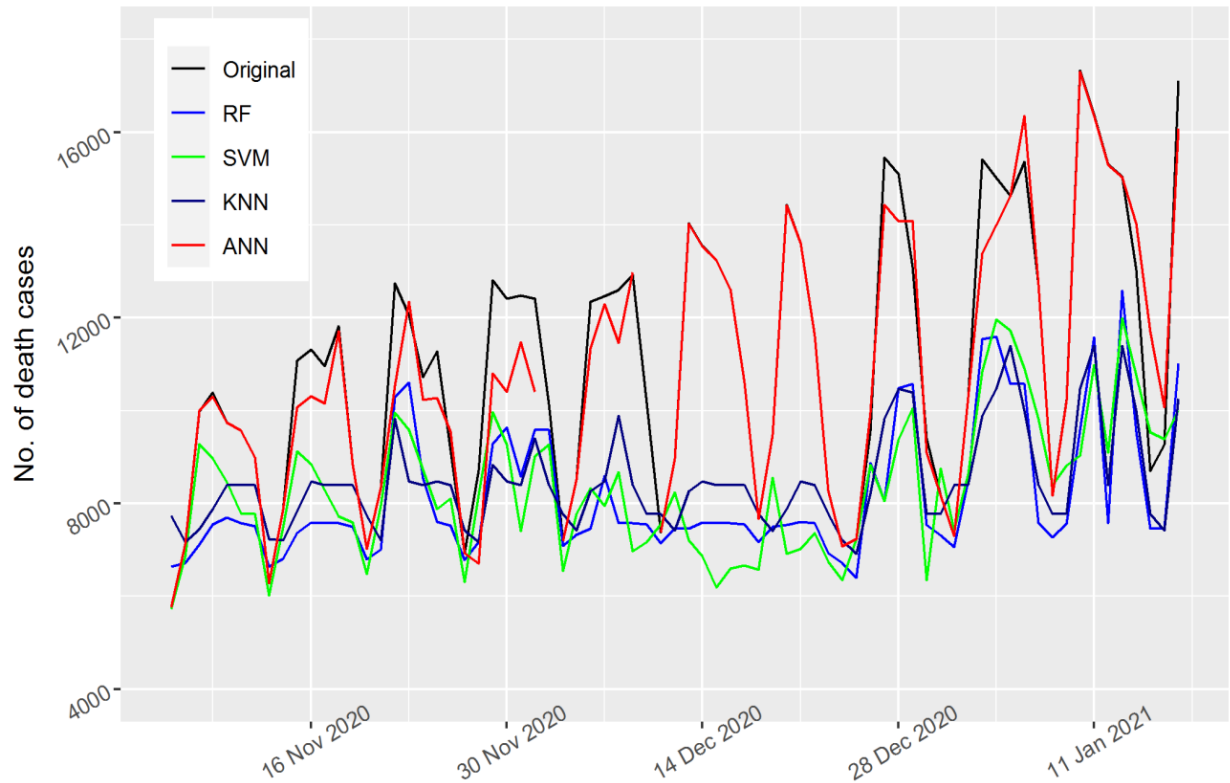

**Fig. 5.** Original and forecasted values of RF, SVM, KNN, and ANN models for daily death cases of COVID-19 of the testing set. The testing set consist of 20% of the daily deaths a total of 74 observations spanning from 08 Nov 2020, to 18 Jan 2021 to validate the models performance. From the plot it is observed that the ANN model forecasted values are very close to original values and also follow the same trend

Fig. 5 displays the prediction accuracy of RF, SVM, KNN, and ANN models. All the models are capable of capturing competently the pattern of the daily death cases of COVID-19. Fig. 5 clearly shows that ANN captured the pattern of the test set of the data better than RF, SVM, and KNN methods. Also, (Fig. 5) displays the prediction accuracy of RF, SVM, KNN, and ANN models for COVID-19's daily recover cases. Similar to death cases accuracy results, all the models effectively captured the pattern of the daily recover cases of COVID-19. In the same way, in Fig. 6 and Fig.7, the ANN captured the pattern on the test part of the data. While the rest of the methods first follow the pattern up to some extent and then insensitive to the original data. The Fig. 6 and Fig. 7 are shown below.

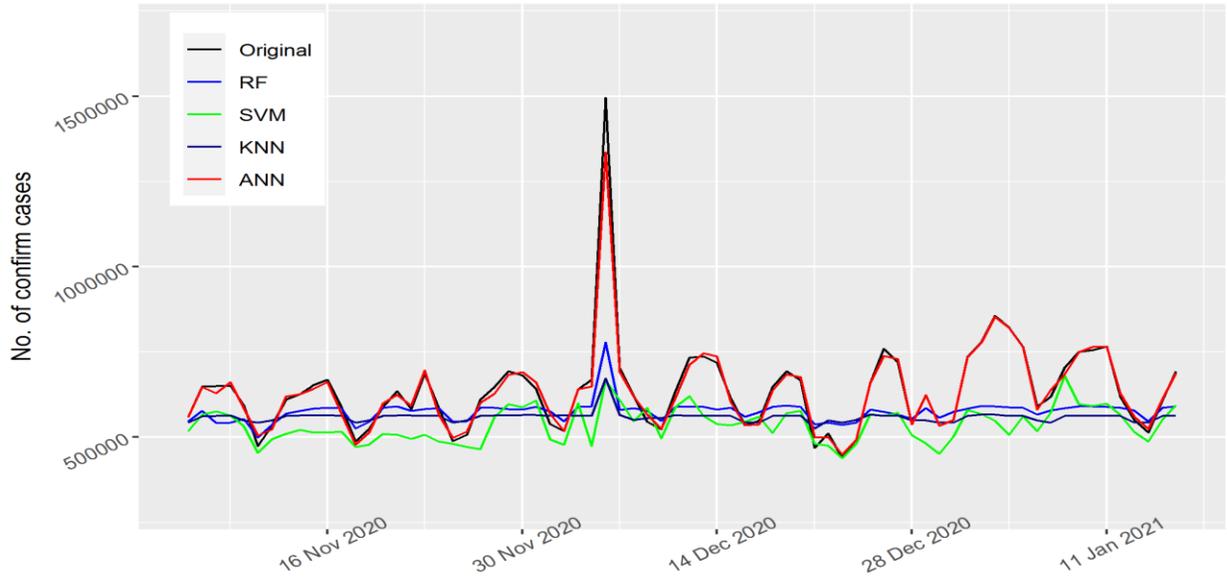

**Fig. 6.** Original and forecasted values of RF, SVM, KNN, and ANN models for confirmed cases of COVID-19 of the testing set. The testing set consist of 20% of the daily confirms a total of 74 observations spanning from 08 Nov 2020, to 18 Jan 2021 to validate the models performance. From the plot it is observed that the ANN model forecasted values are very close to original values and also follow the same trend

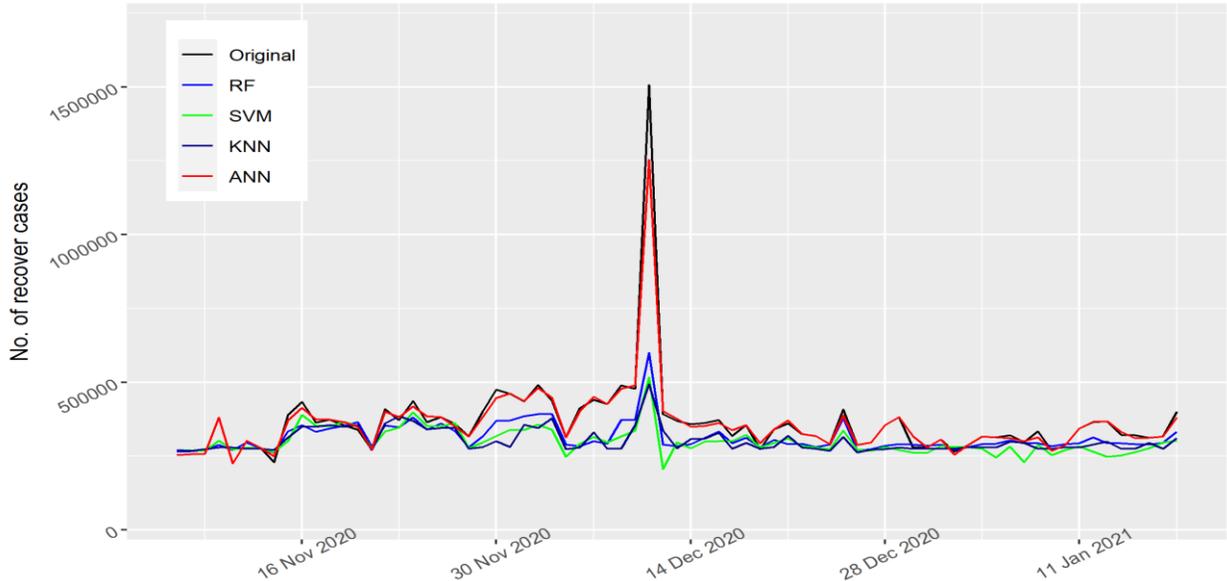

**Fig. 7.** Original and forecasted values of RF, SVM, KNN, and ANN models for recover cases of COVID-19 of the testing set. The testing set consist of 20% of the daily recovered cases and a total of 74 observations spanning from 08 Nov 2020, to 18 Jan 2020 to validate the models performance. From the plot it is observed that the ANN model forecasted values are very close to original values and also follow the same trend

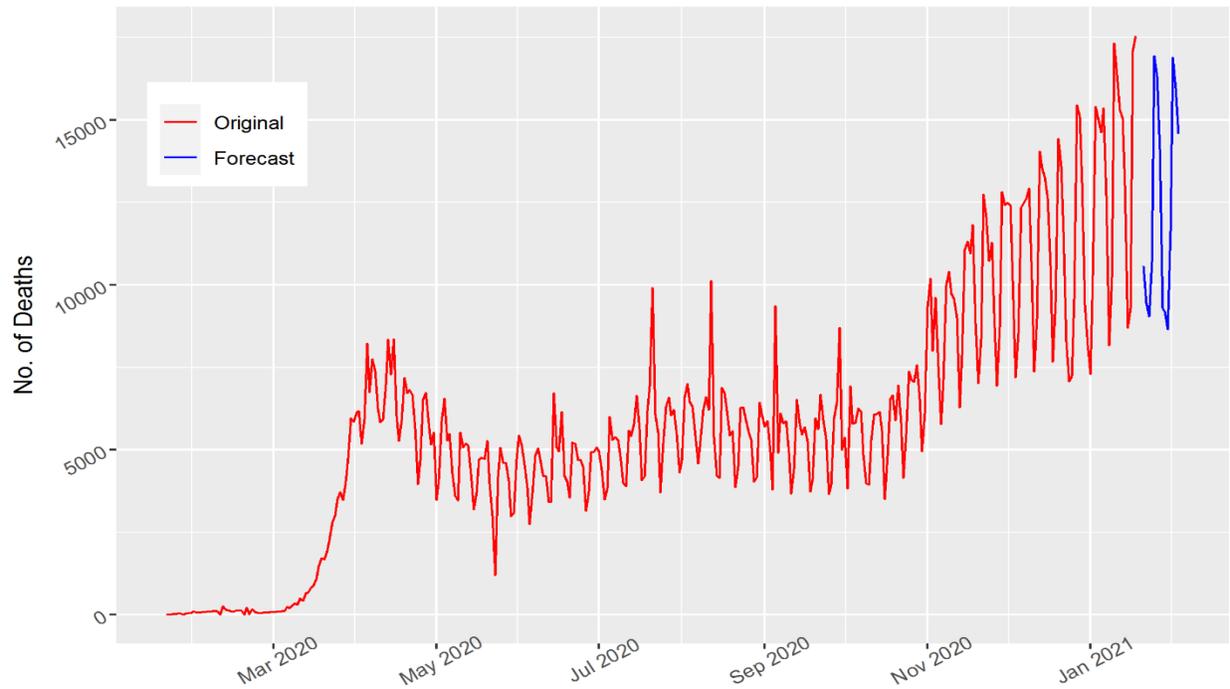

**Fig. 8.** Plot of the original and forecasted values of ANN model for daily deaths cases of Covid-19. The red dots shows the 15 days ahead forecasts spanning from 19 Jan 2021 to 03 Feb 2021. The forecasted number of deaths tend to gradually decline over time. This is an indication that number of daily deaths decreases over time.

In (Fig. 8), the original COVID-19 number of deaths data points and the resulting forecast of ANN were plotted for the next fifteen days from (19 Jan 2021 to 03 Feb 2021). As shown in the figure, the ANN forecast captures and follows the pattern of the original death cases of COVID-19. The subsequent fifteen days forecasted line fluctuated near 10,000. In addition, the forecasted number of deaths tend to gradually decline over time. This is an indication that number of daily deaths decreases over time.

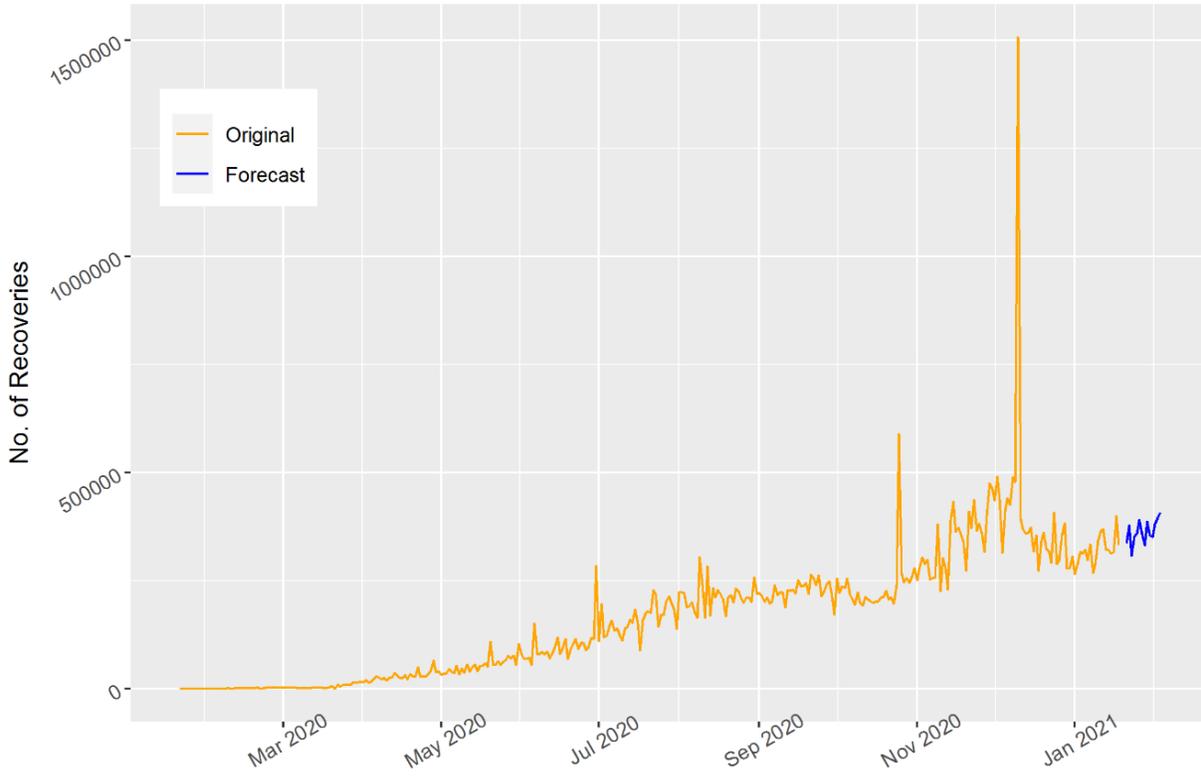

**Fig. 9.** Plot of original and forecasted values of ANN model for daily recovered cases of COVID-19. The red dots shows the 15 days ahead forecasts spanning 19 Jan 2021 to 03 Feb 2021. The forecasted drift going in downward direction. This reveals that the number of daily recoveries is decreasing over time.

In (Fig. 9), the original COVID-19 recover patients data and forecast of ANN exhibited for the next fifteen days from (04Dec 2020 to 18 Dec 2020). The ANN model forecast captured the pattern of the original COVID-19 recover patient's data. In addition, the next fifteen days forecasted drift going in downward direction. This reveals that the number of daily recoveries is decreasing over time.

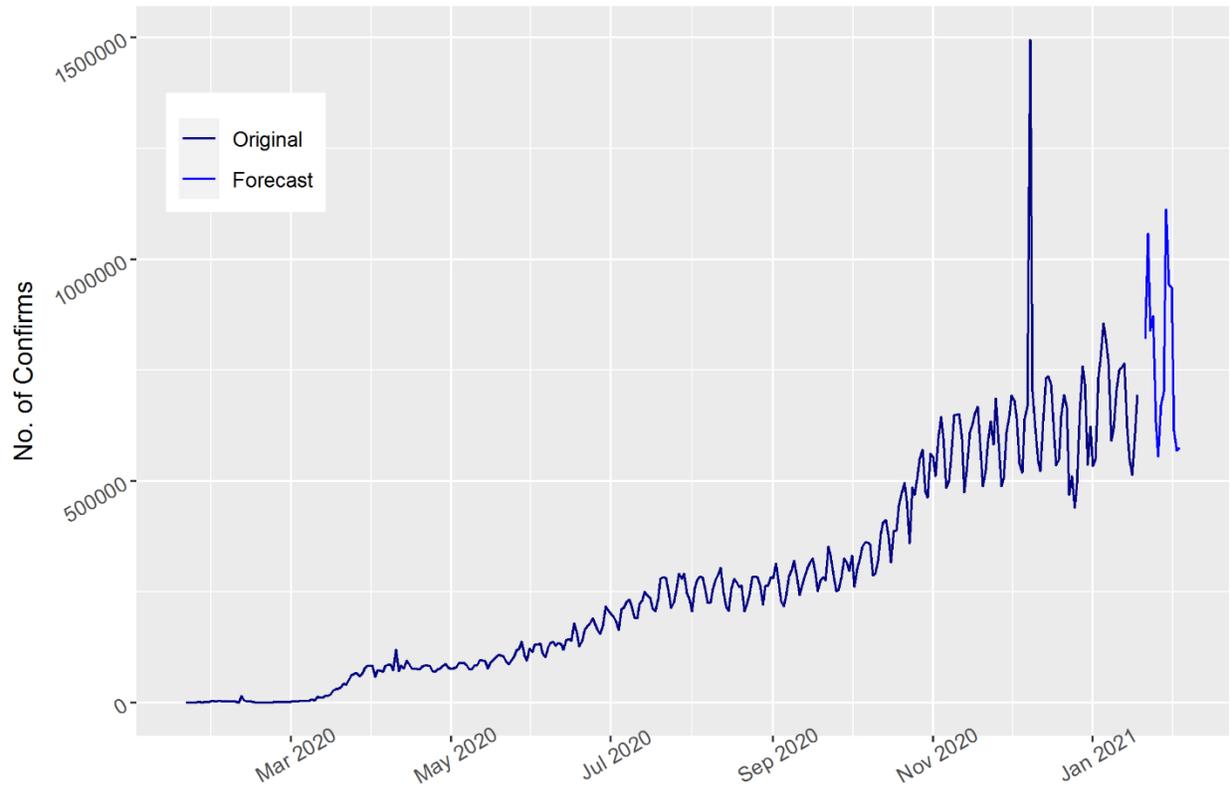

**Fig. 10.** Plot of the original and forecasted values of ANN model for daily confirm cases of COVID-19. The red dots shows the 15 days ahead forecasts spanning from 19 Jan 2021 to 03 Feb 2021. The forecasted drift going in downward direction. This reveals that the number of daily confirm cases is decreasing over time.

In (Fig. 10), the original COVID-19 confirm patients data and forecast of ANN exhibited for the next fifteen days from (19 Jan 2021 to 03 Feb 2021). The ANN model forecast captured the pattern of the original COVID-19 confirm patient's data. In addition, the next fifteen days forecasted drift going in downward direction. This reveals that the number of daily confirm is decreasing over time.

## Conclusions

This paper proposed four predicting models for COVID-19 outbreak. The methods are compared with respect to five performance metrics including ME, RMSE, MAE, MPE, and MAPE. The results for the daily deaths cases are based on 80% training and 20% testing parts. Among the four methods using these performance metrics, the ANN achieved better results in every aspect. In the same way, the results obtained for the daily recovered cases using 80% training and 20% testing parts and ANN have attained better results with respect to the other methods. Moreover, daily confirm cases results obtained using the same training and testing parts and in most of the cases ANN performed better than the other methods. Therefore, the major findings of this study reveal that ANNs outperform the rest of the methods for both models. In addition, ANN suggests consistent prediction performance compared to RF, SVM, and KNN models and hence preferable as a robust forecast model. The AI-based method's accuracy for predicting the trajectory of the COVID-19 is high. For this specific application in predicting the disease, the authors consider the results are reliable. In this study, ANN generates the fastest convergence and good forecast ability in most cases. The results showed the compensations of machine learning algorithms to support strategy/decision-makers in evolving short term policies about the number of disease prevalence. The forecast models will support the government and health staff to be ready for the forthcoming circumstances and take further promptness in healthcare structures. The forecasted figures were calculated for the next fifteen days (i.e., 19 Jan 2021 to 03 Feb 2021) for COVID-19 data.

It is worth noting that forecasting is a complex matter, and some tailored models might not be ubiquitous owing to the complex societal and economic circumstances of different nations. The models and predictions proposed in this article do not reflect the local demography, and the real statistics can variate owing to numerous governmental actions like concentration on lockdown, the strategy of isolation and health facilities, etc. Thus, readers should be careful while interpreting these forecasts.

## Disclosures

No conflicts of interest, financial or otherwise, are declared by the authors.